\documentclass[10pt,twocolumn,letterpaper]{article}
    
\usepackage{color}

\usepackage{cvpr}
\usepackage{url}
\usepackage[T1]{fontenc}
\usepackage{newtxtext,newtxmath}
\usepackage{graphicx}
\usepackage{amsmath}
\usepackage{booktabs}
\usepackage{multirow}
\usepackage{subcaption}
\usepackage{caption}
\usepackage{array}
\newcolumntype{M}[1]{>{\centering\arraybackslash}m{#1}}
\newcolumntype{L}[1]{>{\arraybackslash}m{#1}}

\renewcommand{\arraystretch}{1.2}

\usepackage[pagebackref=true,breaklinks=true,letterpaper=true,colorlinks,bookmarks=false]{hyperref}

\cvprfinalcopy

\def\cvprPaperID{482} 

\ifcvprfinal\fi
\begin{document}

\title{Self-Supervised Learning via multi-Transformation Classification \\ for Action Recognition}


\author{
Duc-Quang Vu$^{1}$ \quad
Ngan Le$^{2}$ \quad
Jia-Ching Wang$^{1}$ \\
$^{1}$Dept. of CSIE, National Central University, Taoyuan, Taiwan \\\
$^{2}$Dept. of CSCE, University of Arkansas, Fayetteville, USA\\
}



\maketitle

\begin{abstract}
  Self-supervised tasks have been utilized to build useful representations that can be used in downstream tasks when the annotation is unavailable. In this paper, we introduce a self-supervised video representation learning method based on the \textbf{multi-transformation classification} to efficiently \textbf{classify human actions}. Self-supervised learning on various transformations not only provides richer contextual information but also enables the visual representation more robust to the transforms. The spatio-temporal representation of the video is learned in a self-supervised manner by classifying \textbf{seven different transformations} i.e. rotation, clip inversion, permutation, split, join transformation, color switch, frame replacement, noise addition. First, seven different video transformations are applied to video clips. Then the 3D convolutional neural networks are utilized to extract features for clips and these features are processed to classify the pseudo-labels. We use the learned models in pretext tasks as the pre-trained models and fine-tune them to recognize human actions in the downstream task. We have conducted the experiments on UCF101 and HMDB51 datasets together with C3D and 3D Resnet-18 as backbone networks. The experimental results have shown that our proposed framework is outperformed other SOTA self-supervised action recognition approaches. The code will be made publicly available.
\end{abstract}

\section{Introduction}
\label{sec:intro}


Human action recognition is one of the most fundamental research problems in computer vision and machine learning. It has attracted huge attention over the last decade with the availability of large-scale video datasets~\cite{i3d_2017, fernando2017self}. However, annotating new video datasets are always required to address the problems in new domains. Annotation is time-consuming and labour-intensive; thus, it is useful if we can leverage the unlabeled data. As one of the most widely used datasets for pre-training very deep 2D CNNS, ImageNet~\cite{deng2009imagenet} contains about 1.3 million labelled images covering 1,000 classes. Furthermore, the collection and annotation of video datasets are more expensive than image datasets due to the temporal dimension. The Kinetics dataset~\cite{carreira2018short}, which consists of 500,000 10-second videos belonging to 600 categories, is mainly used to train network for video action recognition. 

To avoid time-consuming and expensive data annotations, many self-supervised methods~\cite{wang2019self, fernando2017self, ahsan2019video, misra2016shuffle} have been proposed to learn visual features from large-scale datasets for video recognition tasks based on self-supervised learning without using any human annotations. Self-supervised learning constructs a pre-training or “pretext” task used to extract knowledge from unlabeled data. After training a model on the pretext task, it can then be adapted to the target task through transfer learning. Self-supervised learning approaches usually involve transforming the input data to force the model to predict missing parts of the data or recognize the transformations applied to the data or introduce some information bottleneck. A pretext task with pseudo-labels is automatically generated to exploit data structure. A CNN model is then trained to solve tasks where pseudo-labels can be easily derived from input data without human labours. Some examples for those such tasks are solving a jigsaw puzzle of image patches~\cite{ahsan2019video}, predicting frames order~\cite{el2019skip, xu2019self, misra2016shuffle}, motion and appearance statistics~\cite{wang2019self}, image colour channel~\cite{zhang2016colorful}, etc. The CNN can then be directly applied to other video tasks as a feature extractor or to be used as a weight initialization for downstream tasks.


In this paper, we propose a novel self-supervised learning approach to learn video representations by classifying the transformation that was applied to the video. The success of multi-task self-supervised learning inspires our proposed method~\cite{doersch2017multi, sarkar2020self, ravanelli2020multi}. Our proposed framework contains two parts corresponding to pretext task and downstream task, as shown in Fig.\ref{fig:overview_model}. The 3D CNN model is learned in the pretext task by applying seven transformations to change the appearance and/or motion in the input video clip. The transformation is then classified using the pseudo-labels where each label is assigned to one transformation. The 3D CNN model is later used as a pre-trained model for the downstream task to recognize human actions.

We summarize our contributions as follows:
\begin{itemize}
    \item We have proposed an effective self-supervised framework for action recognition based on the multi transformation classification. Different from the existing methods which focus on either spatial or temporal domain, our transformer is able to cover both temporal and spatial domains. In the pretext task, the multi-transformation is applied to the input videos to model the spatio-temporal features.
    
    \item From the ablation study results, we demonstrate that multi-transformations are the motivator to help the model learn more spatio-temporal features from the input clip. Besides, multi-transformations are also a simple and effective data augmentation strategy that compares to a single transformation.
    
    \item We have conducted experiments on C3D and 3D ResNet-18 backbone networks. On both benchmarks, our proposed framework robustly exhibits strong performances i.e. outperforms SOTA approaches on the UCF101 and HMDB51 datasets regardless of the backbone network. 
\end{itemize}


The remainder of the paper is organized as follows. Section~\ref{sec:related_work} provides a review of related work. Our model is proposed in Section~\ref{sec:model}, which includes the pretext and downstream tasks. The loss function for the network is also presented in this section. The experimental results, comparisons, and component analysis are presented in Section~\ref{sec:experiment}. The conclusions of the paper are given in Section~\ref{sec:conclusion}.

\section{Related Work}
\label{sec:related_work}
In this section, we first introduce the recent progress in action recognition. We then discuss recent work on self-supervised representation learning.

\subsection{Action Recognition}

Action recognition has always been one of the most important topics in computer vision. The traditional methods proposed to solve this problem are based on efficient spatio-temporal feature representations and motion propagation across frames in videos such as HOG3D~\cite{klaser2008spatio}, SIFT3D~\cite{scovanner20073}, ESURF~\cite{willems2008efficient}, MBH~\cite{dalal2006human}, iDTs~\cite{wang2013action}, etc.

Following the success of CNN on image tasks. Tran et al.~\cite{tran2015learning} proposed a simple model named C3D which outperforms all previous best-reported methods. By transferring the 2D pre-trained model to 3D model, Carreira et al.~\cite{i3d_2017} proposed I3D. In I3D, the 3D filters are replaced by a set of repeated 2D filters. Inspired by the success of ResNet in image classification, Hara et al.~\cite{resnet3D_50} extended ResNet architecture to 3D CNN and proposed 3D ResNet. In their work, they examined various 3D CNN architecture under different backbone such as ResNet-18, ResNet-34, ResNet-50, ResNet-101, ResNet-152, ResNet-200, DenseNet-121 and ResNeXt-101. 

Achieving state-of-the-art performance is the most important task in action recognition. Most methods adopt the ResNet backbone as a standard architecture to modify. The STM network was introduced in~\cite{jiang2019stm}. In this network, the authors presented a method to enhance the ability to learn Spatio-temporal and motion features from a video. To do that, the authors proposed to encode these two features in a unified 2D framework. The Channel-wise SpatioTemporal Module (CSTM) is used to learn spatiotemporal features and the Channel-wise Motion Module (CMM) is for encoding motion features. These two modules are added to the original residual blocks in the ResNet architecture. The experiment shows that STM performs a little better than major proposed 3D CNN and 2D CNN based methods. Inspired by the evolutionary algorithms in the optimization field such as the genetic algorithm, Piergiovanni et al.~\cite{piergiovanni2019evolving} proposed a new method for finding video CNN architectures. In their work, a novel evolutionary search algorithm is developed to automatically explores different types of models and combine layers based on mutation operations. And then they obtain new architectures superior to manually designed architectures. There are three mutation operations proposed in this paper including "Change Layer", "Change Temporal Size" and "Add Layer". However, the crossover operator is not mentioned in this paper. To find good architectures with state-of-the-art performance, the authors built a population with 2000 different CNN architectures, and each newly generated child architecture (from their parent) is trained for 1000 iterations. Due to training 3D CNNs being computationally expensive, so 12 GPUs are used for this task. 


\subsection{Self-Supervised Representation Learning}

\begin{figure*}
  \centering
  \centerline{\includegraphics[scale=0.49]{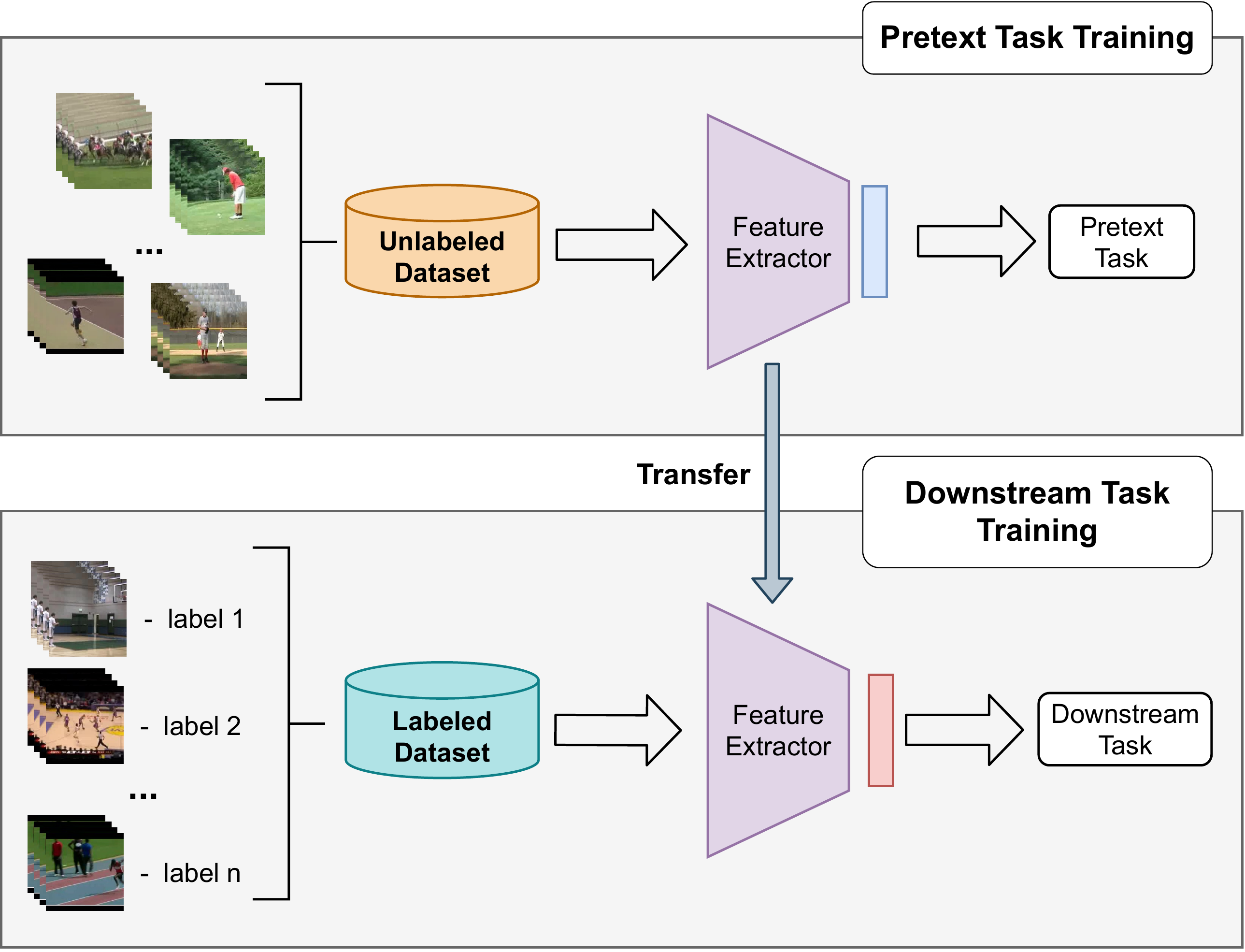}}
  \caption{Overview of a self-supervised learning model for action recognition. This process consists of two phases, including pretext task training and downstream task training.}\medskip
  \label{fig:overview_self_supervised}
\end{figure*}

Self-supervised learning aims at learning visual features from unlabeled data in pretext tasks. The learned visual representation model in the pretext task is then transferred to the downstream task. The objective of self-supervised learning focuses on extracting good feature representations without annotation; thus, it targets designing an effective pretext task component. 

Villegas et al.~\cite{villegas2017decomposing} proposed a deep neural network that uses both optical flow frames and the RGB frame to predict one future frame. With the input as a tuple of frames order, Misra et al.~\cite{misra2016shuffle} proposed a method that allows verifying whether the temporal order is correct or not. To solve this pretext task, the authors proposed a ConvNet model that all input video frames are passed through the model. The objective of the model figures out whether the frames are in the correct order or not. In doing so, the model learns not just spatial features but also takes into account temporal features. 

Inspired by the frames reordering task, Kim et al.~\cite{kim2019self} introduced a self-supervised task called Space-Time cubic puzzles. Given a randomly permuted sequence of 3D spatio-temporal pieces cropped from a video clip.  The 3D CNN is used to learn both spatial and temporal relations from the input video frames and predict their original arrangement. Fernando et al.~\cite{fernando2017self} presented a self-supervised CNN called O3N to predict an odd video from a set of otherwise related input videos. The goal of O3N is to predict an odd video from a set of otherwise related input videos. The network's input is a tuple of videos where one of the videos has the wrong temporal order of frames while the other ones have the correct temporal order. 

Unlike the above methods, a model based on deep reinforcement learning is introduced in~\cite{buchler2018improving}. In this work, the deep reinforcement model is proposed to learn a policy that proposes best-suited permutations from errors the model has made when recovering frame order. Wang et al.~\cite{wang2019self} presented a pretext task to predict motion and appearance statistics. Each video frame is first divided into several spatial regions, and it then predicted by selecting the largest motion and direction.

\begin{figure*}
  \centering
  \centerline{\includegraphics[scale=0.8]{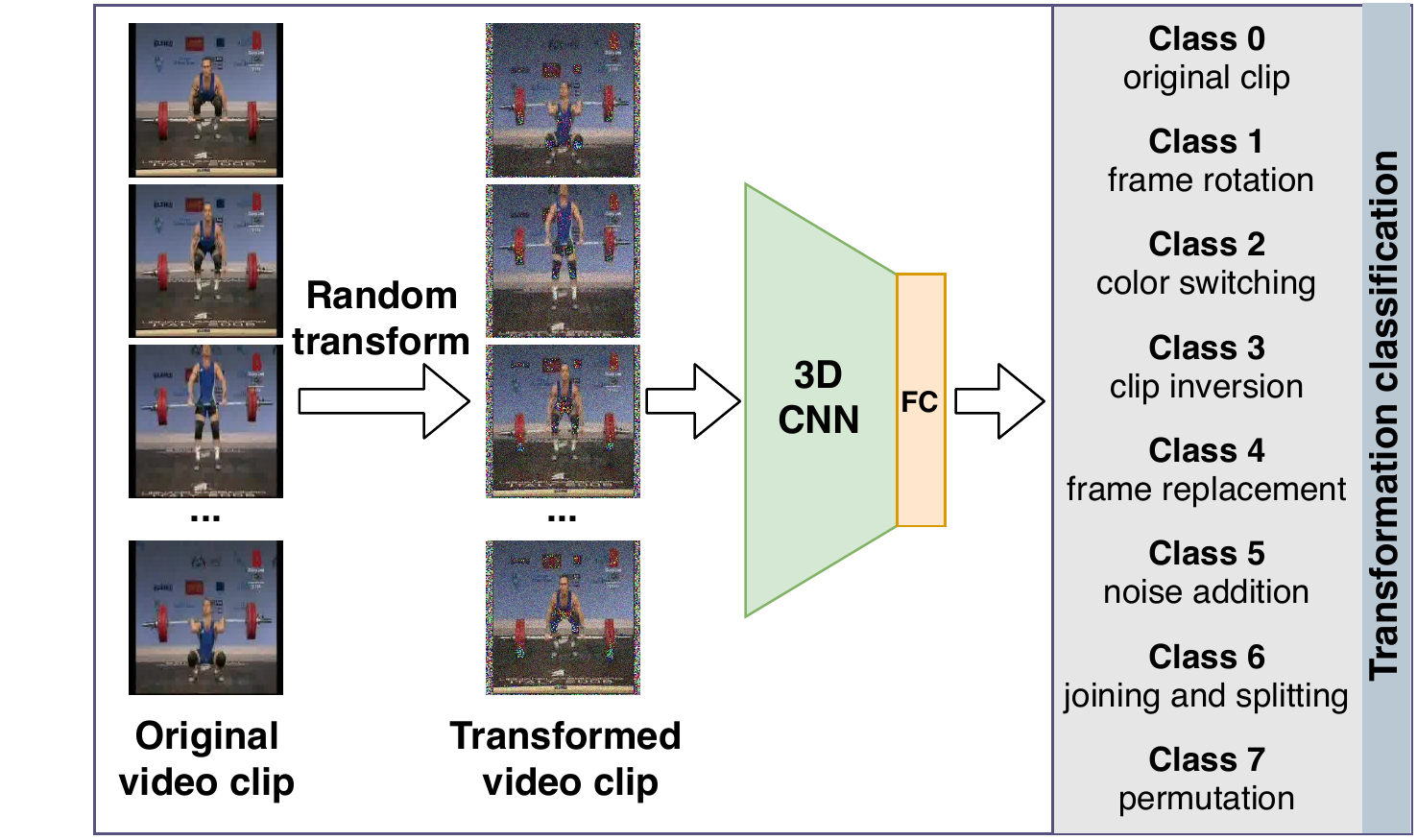}}
  \caption{Overview of self-supervised learning via multi-transformation classification.}\medskip
  \label{fig:overview_model}
\end{figure*}

Far apart from the previous self-supervised learning approaches, which focus only on single transformation during training pretext task, our proposed network inherits the advantages of multiple transformations, including frame rotation, color switching, video clip inversion, noise addition to frames, video joining and splitting, permutation, and frame replacement. Learning various transformations not only provides richer contextual information but also enables the visual representation to be more robust to the transformations.

\section{Proposed Model}
\label{sec:model}


Generally, a self-supervised learning framework contains modules corresponding to two tasks: pretext task (PT) and downstream task (DT). First, we build a 3D CNN to learn video representation by solving the PT. We then transfer the network capable of extracting video representation for DT, i.e. action recognition, in this work. The general pipeline of self-supervised learning for action recognition is shown in Fig.~\ref{fig:overview_self_supervised}

\subsection{Pretext Task}


Let $x' = \mathcal{G}(x,m)$ be the transformed video where $x$ denotes the original video and $m$ indicates the label of transformations. In this task, we adopt a 3D CNN $\mathcal{F}(x'|\theta)$ to learn the spatio-temporal features by predicting $m$, and $\theta$ is the set of trainable parameters. Given a video $x_i$, $\theta$ is learned by minimizing the following objective function:

\begin{equation}
\begin{split}
    \mathcal{L}(x_i|\theta) = - \sum_{z=0}^M (z \log \mathcal{F}(x'_i|\theta) + (1-z) \log(1-\mathcal{F}(x'_i|\theta))
\end{split}
\label{eq:loss_m}
\end{equation}

where $M$ corresponds to the total number of transformations. In our approach, we randomly select a few transforms among M (M = 7) transforms and apply them into one video.  As shown in Eq~\ref{eq:loss_m}, we use the cross-entropy loss for the binary classification to predict if a single transformation $m$ is applied to a video or not. After examining $M$ transformations, we receive a vector in M-dimensional space. The $m^{th}$ element in the vector presents the probability of applying transformation $m$. Given a set of $N$ training videos $\{x_{i}\}_{i=0}^{N}$, the overall training loss function is defined as:

\begin{equation}
    \mathcal{L}_{\mathrm{PT}} = \min_{\theta} \frac{1}{N} \sum_{i=0}^N \mathcal{L}(x_i|\theta)
\end{equation}

The overview of our proposal is given in Fig.~\ref{fig:overview_model}. We propose seven different transformations as follows:

\noindent 
\textit{\textbf{Frame rotation}}: Different rotation angles i.e.  $90^o$, $180^o$, or $270^o$ are applied into frames.\\
\noindent 
\textit{\textbf{Color switching}}: Besides the actual color channel order as R-G-B, the color channel is inverted to other orders such as B-G-R.\\
\noindent 
\textit{\textbf{Noise addition}}: In this transformation, a Gaussian distribution with zero mean, and standard deviation is randomly set from 0.1 to 0.3 is added into frames. \\
\noindent 
\textit{\textbf{Frame replacement}}: In this transformation, a frame from the given video clip is replaced by a noise frame that is generated from a uniform distribution. \\
\noindent 
\textit{\textbf{Clip inversion}}: Given an original video clip $X(t)$ where $t=1,2,...,N$ and $N$ is the length of the video clip, the inverted clip is expressed as $X'(t)$ where $t=N, N-1, ...,1$. \\
\noindent 
\textit{\textbf{Splitting and joining}}: In this transformation, a video clip is divided into two parts and one of them is replaced by another part from another video clip which has the same length/dimension \\
\noindent 
\textit{\textbf{Permutation}}: All frames in the given clip are randomly shuffled.

\noindent 

We apply seven transformations to the video clip and send these transformed images to the pretext task network to predict what sort of transformation was applied to the video clip and the network simply performs an 8-class classification to predict the transformation.

\begin{figure}[!ht]
  \centering
  \centerline{\includegraphics[scale=0.36]{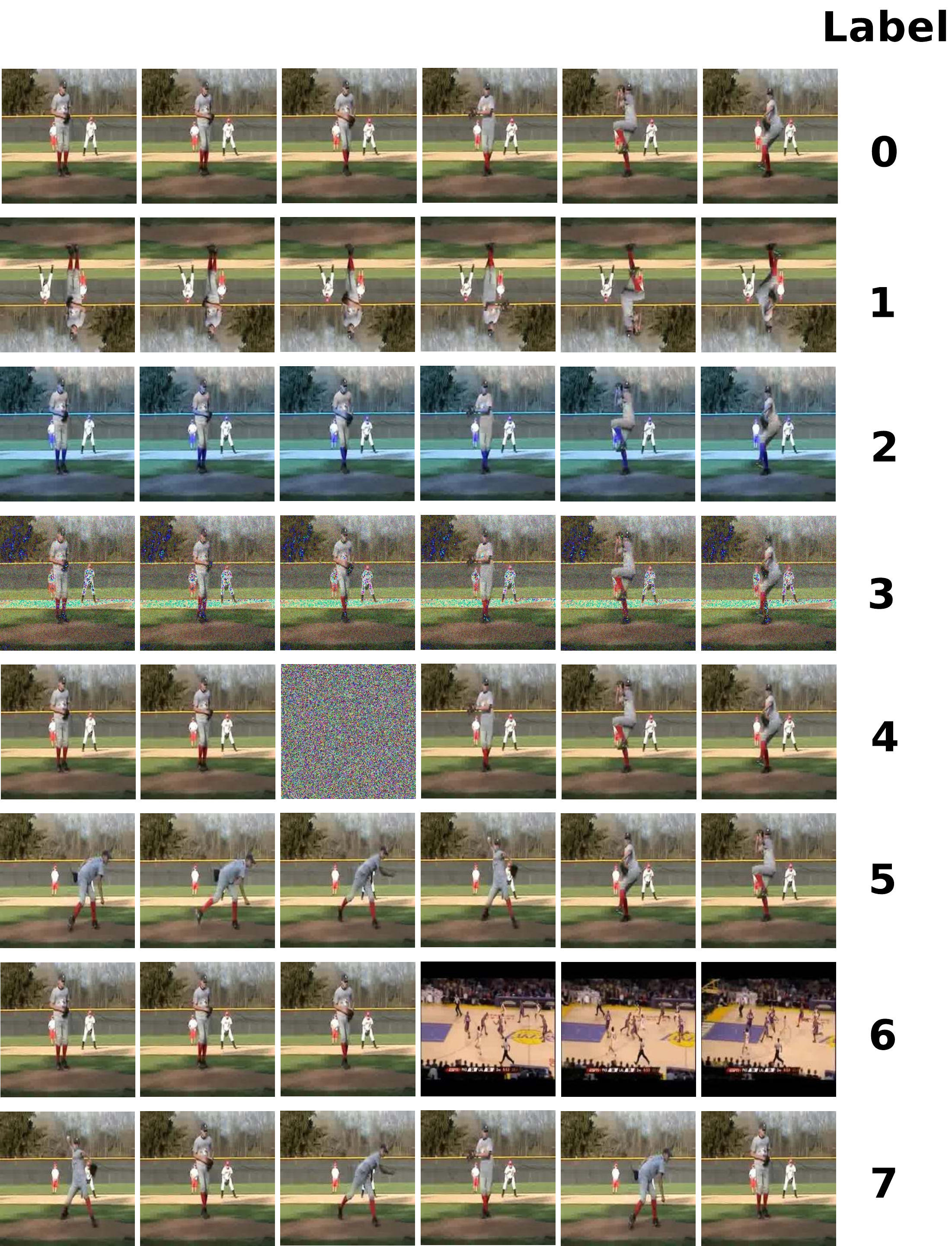}}
  \caption{An illustration of seven different transformations that are applied to the original video clip given in the $1^{st}$ row. Each transformation is assigned one pseudo label. Label 0: original video clip, label 1: frame rotation, label 2: color switch, label 3: noise addition, label 4: frame replacement, label 5: clip inversion, label 6: split and join, label 7: frames permutation.}\medskip
  
  \label{fig:example_trainsformations}
\end{figure}

Fig.\ref{fig:example_trainsformations} illustrates examples of transformations together with the pseudo labels. The original video clip is given in the first row, and its pseudo label is zero. In contrast, the remaining rows are transformed video clips, and their pseudo labels are assigned from 1 to 7, corresponding to different transformations. 

\subsection{Downstream Task}

In our self-supervised learning framework, the downstream task performs action recognition to evaluate the quality of features learned by the pretext task. Given a set of $N$ samples is denote as ${(x_1, y_1), (x_2, y_2), ..., (x_N, y_N)}$ such that $x_i$ is a video clip and $y_i$ is its label (i.e., action). In this stage, the model $\mathcal{F}$ is fine-tuned to learn a mapping $\mathcal{F}: x_i \rightarrow y_i$. Let $p_i$ be a probability distribution over labels, where $p_i = \mathcal{F}(x_i , \theta')$ and $\theta'$ is the set of trainable parameters. The correctness of the prediction is measured using cross-entropy as follows:

\begin{equation}
    \mathcal{L}_{\mathrm{DT}}(y_i, p_i) = - \frac{1}{N} \sum_{i=0}^N y_i \log(p_i)
\end{equation}

The loss $\mathcal{L}_{\mathrm{DT}}$ is then backpropagated to optimize the whole framework. When the model is trained to predict the action classes, the 3D CNN is trained to extract clips' meaningful features. A fully connected layer with the softmax function is applied over to output the final prediction.

\subsection{Network Architecture}
In our work, we consider two state-of-the-art convolutional neural network architectures: C3D~\cite{tran2015learning} and 3D ResNet-18~\cite{resnet3D_50}. 

C3D is a natural extension of 2D CNNs on videos. Far apart from 2D CNNs, 3D CNNs can learn temporal information; thus, they are well-suited for videos. The C3D network includes eight 3D convolution layers interleaved with five pooling layers and followed by two fully connected layers. The C3D network is described in Table~\ref{tab:c3d_architecture}.

 \begin{table}[!ht]
 \centering
  \caption{The C3D network architecture. The FC is a fully connected layer. The Pool means a max-pooling layer, and the Conv is a convolution block that includes a convolution layer and followed by Batch Normalization and ReLU layers.}\medskip
  \setlength\tabcolsep{6pt}
  \renewcommand{\arraystretch}{1.5}
\begin{tabular}{c|c|c}
\specialrule{1pt}{0pt}{1pt}
\specialrule{1pt}{0pt}{1pt}
\textbf{Layer}  & \textbf{Specification} & \textbf{Output size}    \\
\specialrule{1pt}{0pt}{1pt}
\specialrule{1pt}{0pt}{1pt}
Input       &                & $T \times 224 \times 224 \times3$  \\
\hline
Conv\_1       &  $7\times7\times7, 64$ & $T \times 224 \times 224 \times 64$ \\
\hline
Pool\_1        & stride = $1 \times 2 \times 2$  & $T\times112\times112\times64$   \\
\hline

Conv\_2   &  $3 \times 3 \times 3, 128 $   & $T \times 112 \times 112 \times 128$   \\
\hline

Pool\_2        & stride = $2 \times 2 \times 2$  & $\frac{T}{2} \times56\times56\times128$   \\
\hline

Conv\_3a   &  $3 \times 3 \times 3, 256 $   & $\frac{T}{2} \times 56 \times 56 \times 256$   \\
\hline

Conv\_3b   &  $3 \times 3 \times 3, 256 $   & $\frac{T}{2} \times 56 \times 56 \times 256$   \\
\hline

Pool\_3        & stride = $2 \times 2 \times 2$  & $\frac{T}{4} \times 28 \times 28 \times 256$   \\
\hline

Conv\_4a   &  $3 \times 3 \times 3, 512 $   & $\frac{T}{4} \times 28 \times 28 \times 512$   \\
\hline

Conv\_4b   &  $3 \times 3 \times 3, 512 $   & $\frac{T}{4} \times 28 \times 28 \times 512$   \\
\hline

Pool\_4        & stride = $2 \times 2 \times 2$  & $\frac{T}{8} \times 14 \times 14 \times 512$   \\
\hline

Conv\_5a   &  $3 \times 3 \times 3, 512 $   & $\frac{T}{8} \times 14 \times 14 \times 512$   \\
\hline

Conv\_5b   &  $3 \times 3 \times 3, 512 $   & $\frac{T}{8} \times 14 \times 14 \times 512$   \\
\hline

Pool\_5        & stride = $2 \times 2 \times 2$  & $\frac{T}{16} \times 7 \times 7 \times 512$   \\
\hline

FC\_1     &     & 4096    \\ \hline
FC\_2     &     & 4096    \\ \hline
FC\_3     & $M$ classes    & $M$    \\
\specialrule{1pt}{0pt}{1pt}
\specialrule{1pt}{0pt}{1pt}
\end{tabular}
\label{tab:c3d_architecture}
\end{table}

The skip connection technique is introduced in~\cite{he2016deep}, ResNet has quickly become the most popular architecture in computer vision. ResNet significantly increases the performance of many image-related tasks such as classification, detection, and segmentation. 3D ResNet-18 (R3D) is an extension of ResNet-18 over videos. In 3D ResNet-18, there are five 3D convolution blocks where each block includes two 3D convolutions, with batch normalization and ReLU layers appended. The 3D ResNet-18 is described in Table~\ref{tab:res18_architecture}.

 \begin{table}[!ht]
 \centering
  \caption{The 3D ResNet-18 network architecture. In which, the FC is a fully connected layer, the Pool means a max pooling layer. Each convolution layer is followed by a Batch Normalization and a ReLU layer. Spatial AVE Pool means a average pooling layer that is calculated along the spatial dimension. }\medskip
  \setlength\tabcolsep{1.5pt}
  \renewcommand{\arraystretch}{1.4}
\begin{tabular}{c|c|c}
\specialrule{1pt}{0pt}{1pt}
\specialrule{1pt}{0pt}{1pt}
\textbf{Layer}  & \textbf{Specification} & \textbf{Output size}    \\
\specialrule{1pt}{0pt}{1pt}
\specialrule{1pt}{0pt}{1pt}
Input       &                & $T \times 224 \times 224 \times3$  \\
\hline
 Conv1       & \begin{tabular}[c]{@{}c@{}}$7\times7\times7, 64$\\ stride= $1 \times 2 \times 2$\end{tabular} & $T\times112\times112\times64$ \\
\hline
Pool        & \begin{tabular}[c]{@{}c@{}}$3\times3\times3$\\ stride = $1\times 2\times 2$ \end{tabular}    & $T\times56\times56\times64$   \\
\hline

\rule[-20pt]{0pt}{45pt} Conv block 2   &  $\left[\begin{array}{c} 3 \times 3 \times 3, 64 \\ 3 \times 3 \times 3, 64 \end{array}\right] \times 2$     & $T \times 56 \times 56 \times 64$   \\
\hline

\rule[-20pt]{0pt}{45pt} Conv block 3  & $\left[\begin{array}{c} 3 \times 3 \times 3, 128 \\ 3 \times 3 \times 3, 128 \end{array}\right] \times 2$            & $\frac{T}{2} \times28\times28\times128$   \\
\hline
\rule[-20pt]{0pt}{45pt} Conv block 4  & $\left[\begin{array}{c} 3 \times 3 \times 3, 256 \\ 3 \times 3 \times 3, 256 \end{array}\right] \times 2$            & $\frac{T}{4}\times14\times14\times256$   \\
\hline
\rule[-20pt]{0pt}{45pt} Conv block 5  & $\left[\begin{array}{c} 3 \times 3 \times 3, 512 \\ 3 \times 3 \times 3, 512 \end{array}\right] \times 2$            & $\frac{T}{8}\times7\times7\times512$     \\
\hline
\begin{tabular}[c]{@{}c@{}}Spatial\\ AVE Pool\end{tabular} &          & $\frac{T}{8} \times 1 \times 1 \times 512$         \\
\hline
Flatten     &     &  $\frac{T}{8} \times 512$   \\
\hline
FC     &  $M$ classes   & $M$    \\
\specialrule{1pt}{0pt}{1pt}
\specialrule{1pt}{0pt}{1pt}
\end{tabular}
\label{tab:res18_architecture}
\end{table}

\section{Experiment}
\label{sec:experiment}
\subsection{Datasets and Implementation}
We have conducted experiments on two datasets including HMDB51~\cite{hmdb51} and UCF101~\cite{ucf101}.

\noindent
\textbf{HMDB51:} is a small dataset including 6,766 videos from 51 human action classes. The average duration of each video is about 3 seconds. Three train/test splits (70\% training and 30\% testing) are provided in this dataset.

\noindent
\textbf{UCF101:} is similar to HMDB51. UCF101 includes 13,320 action instances from 101 human action classes. The average duration of each video is about 7 seconds. Three train/test splits (70\% training and 30\% testing) also are provided in this dataset.

\noindent
\textbf{Training the pretext task:} We split the videos in the datasets into many clips with 16 contiguous frames of length. Each frame of the clip is scaled with the shorter edge of 256, and the other edge is calculated so that it still maintains the frame aspect ratio. Then, frames are randomly cropped using window sizes of $224\times224$ (centre cropped for the testing process). To create a transformed video clip from the original, we randomly chose the transformations described above and applied them to the video clip. The pseudo-labels is generated correspondingly for each transformation. We set the mini-batch size to 16. We use stochastic gradient descent (SGD) optimizer with an initial learning rate of 0.01 and a momentum of 0.9. The training process is done in 100 epochs.


\noindent
\textbf{Training the downstream task:} When the pre-training stage with the pretext task is done, we transfer the model to the downstream task. We set the mini-batch size of 16 and the initial learning rate of 0.001. We used SGD optimizer with a momentum of 0.9. Each video is split into different 16-frame clips in the testing, and the class scores are averaged over all the video clips.

\subsection{Performance and Comparison}
To demonstrate the quality of the learned video features from our self-supervised models, we fine-tune our models on the action recognition datasets. As shown in Table~\ref{tab:result}, all results are top-1 accuracy in action recognition on two standard datasets. The results in the table contain three parts. 

\begin{table}[!ht]
\centering
  \caption{Top-1 test accuracy (\%) on UCF101 and HMDB51 datasets. The best performance is shown in \textcolor{blue}{\textbf{blue}} while the second-best is shown in \textbf{bold}. * denotes that these methods use the Kinetics dataset for pre-training.}\medskip
  \setlength\tabcolsep{1.pt}
  \renewcommand{\arraystretch}{1.3}
\begin{tabular}{lccc}
\specialrule{1pt}{0pt}{1pt}
\specialrule{1pt}{0pt}{1pt}
\textbf{Method}               & \textbf{Backbone}     & \textbf{UCF101} & \textbf{HMDB51} \\ 
\specialrule{1pt}{0pt}{1pt}
\specialrule{1pt}{0pt}{1pt}
Random Init          & C3D          & 45.4    & 19.7    \\

\hline
Random Init          & 3D ResNet-18 & 46.5    & 17.1    \\

\specialrule{1pt}{0pt}{1pt}

Shuffle \& Learn \cite{misra2016shuffle}     & AlexNet      & 50.9    & 19.8    \\ \hline
OPN \cite{lee2017unsupervised}   & VGG-M-2048   & 59.8    & 23.8    \\ \hline
O3N - Sum-of-diff. \cite{fernando2017self}    & AlexNet      & 54.3    & 25.9    \\ \hline
O3N - Dynamic image \cite{fernando2017self}   & AlexNet      & 53.2    & 26.0        \\ \hline
O3N - Stack-of-Diff. \cite{fernando2017self}   & AlexNet      & 60.0    & 32.5        \\ \hline
Geometry \cite{gan2018geometry}  & 3D ResNet-18 & 54.1 & 22.6 \\  \hline
CrossLearn \cite{CrossLearn}  & CaffeNet & 58.7 & 27.2 \\ \hline
Video Jigsaw* \cite{ahsan2019video}   & C3D          & 55.4    & 27.0    \\ \hline
Appearance \cite{wang2019self}          & C3D          & 48.6        & 20.3        \\ \hline
Motion \cite{wang2019self}              & C3D          & 57.8        & 29.9        \\ \hline
Motion \& Appearance \cite{wang2019self}  & C3D          & 58.8    & 32.6    \\ \hline
Motion \& Appearance* \cite{wang2019self}  & C3D          & 61.2    & 33.4    \\ \hline
DPC \cite{han2019video}   & 3D ResNet-18 & 60.6    & -       \\ \hline
Geometry \cite{gan2018geometry}            & CaffeNet     & 55.1    & 23.3    \\ \hline
CMC \cite{CMC}  & ResNet-50 & 59.1 & 26.7 \\

\specialrule{1pt}{0pt}{1pt}
      
Ours            & C3D & \textbf{62.8}       & \textbf{34.2} \\ \hline
Ours               & 3D ResNet-18 & \textcolor{blue}{\textbf{63.2}}       & \textcolor{blue}{\textbf{35.9}}
\\
\specialrule{1pt}{0pt}{1pt}
\specialrule{1pt}{0pt}{1pt}
\end{tabular}
\label{tab:result}
\end{table}

+ The first part includes the accuracy of both networks (C3D and 3D ResNet-18) that trained from scratch on the UCF101 and HMDB51 datasets.

+ The second part is shown the performance of state-of-the-art self-supervised methods regardless of backbone networks. All previous methods are pre-trained on different pretext tasks. When transferring to the downstream task (action recognition in this case), these models have different performance that depends on the features the models learned from the pretext task.

+ The third part is our method's performance on C3D and 3D ResNet-18 networks. As can be seen in Table~\ref{tab:result}, the accuracy of both networks is significantly improved compared with training from scratch (increase 16.7\% and 14.5\% on the UCF101 and HMDB51, respectively). This demonstrates that human action recognition can be significantly improved with self-supervised learning regardless of backbone networks, thank to the contextual feature representation learned through the pretext tasks. 

Also in Table~\ref{tab:result}, we compare our results to state-of-the-art self-supervised methods using the RGB video data such as O3N~\cite{fernando2017self}, Video Jigsaw~\cite{ahsan2019video}, Motion \& Appearance*~\cite{wang2019self}, DPC~\cite{han2019video}, Geometry~\cite{gan2018geometry}, and so on. Compare to the existing approaches on various backbone networks and different pretext tasks, our proposed method obtains state-of-the-art performance in terms of accuracy on both datasets. In particular, comparing with several methods that used the Kinetics dataset in the pretext task, we achieve 62.8\% on the UCF101 dataset. This result outperforms Video Jigsaw (\cite{ahsan2019video}) and Motion \& Appearance methods~\cite{wang2019self} by a margin of 2.0\% with the same backbone.


\subsection{Ablation Study}

\begin{figure}[!ht]
  \centering
  \centerline{\includegraphics[scale=0.45]{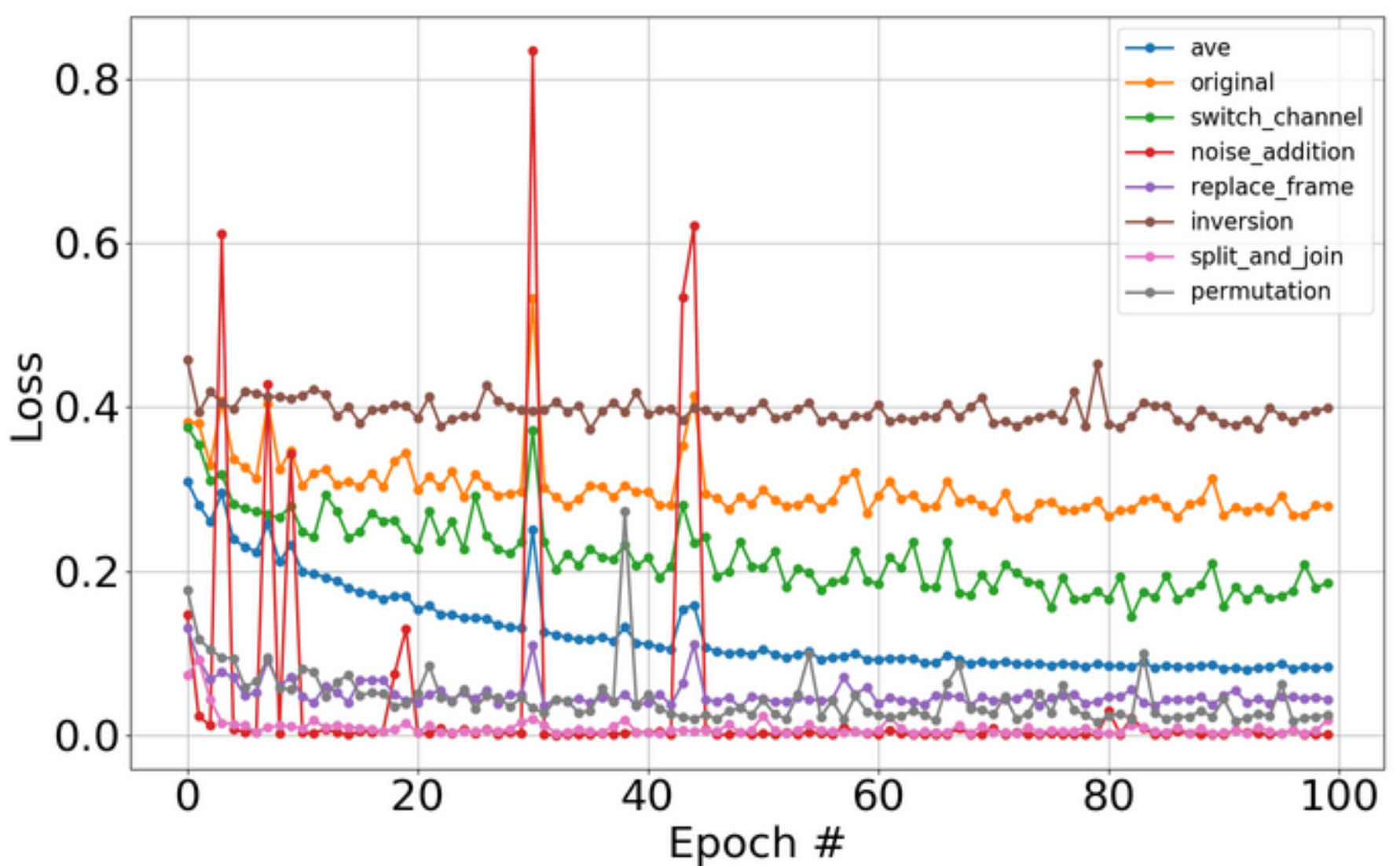}}
  \caption{Individual validation losses for each transformation and average loss (ave) versus epoch are presented for the video transformations recognition task.}\medskip
  \label{fig:ave_loss}
\end{figure}

Fig.~\ref{fig:ave_loss} shows the validation loss for each transformation at 100 epochs in the pretext task. As can be seen in Fig.~\ref{fig:ave_loss}, the transformation recognition network reaches a steady-state when learning the transformation tasks after 80 epochs. However, the different transformations reach their steady states at different epochs. Moreover, clear differences among the steady-state loss of the different transformations are observed, pointing to varied difficulties in the self-supervised training tasks. 

From Fig.~\ref{fig:ave_loss}, we can see that the combination of the proposed transforms aims to generalize the model better. Besides, compared to individual transform, the combination of seven transforms produces much more stable loss during validation.


 \begin{table}[!ht]
\centering
\caption{Evaluation results of each transformation vs multi-transformations on UCF101. All results are top-1 accuracy of action recognition and evaluated on the same C3D network.} 
\medskip
\setlength\tabcolsep{10pt}
  \renewcommand{\arraystretch}{1.2}
  \begin{tabular}{lc}
  \hline
  \hline
  \textbf{Method} & \textbf{UCF101} \\
    \hline
  \hline
  Frame rotation & 52.2 \\
  Color switching & 47.1 \\
  Noise addition & 46.9 \\
  Frame replacement & 47.3 \\
  Clip inversion & 49.6 \\
  Splitting and joining & 46.6 \\
  Permutation & 51.8 \\
  \hline
  \textbf{Multi-Transforms (Ours)} & \textbf{62.8} \\
  \hline
  \hline
  
  \end{tabular}
\label{tab:ablation_study}
\end{table}

To demonstrate the benefit from multi-transformations to action recognition, we conducted the ablation study on each transformation's effectiveness and compared it to multi-transformations. The Table.~\ref{tab:ablation_study} shows top-1 accuracy on the UCF101 dataset. Each transformation is pre-trained with the respective pretext task and then transfer to action recognition. We can see that multi-transformations' performance is significantly improved (increase at least 10.6\%) compared to single-transformation. With more transformations and many transformations being applied to an input video, it is tough for the model to predict which transformations are used. This is the motivator to help the model learn more spatio-temporal features based on transformations. Moreover, many transformations can be proposed for pseudo-label in the pretext task and data augmentation. This is one of the most advantages of the proposed multi-transformation method.

\section{Conclusion}
\label{sec:conclusion}
This paper introduced a novel approach for self-supervised spatio-temporal video representation learning by predicting a set of video transformations. The task is very suitable for 3D CNNs, which can model the spatio-temporal information. From the experimental results, we found that our method achieves state-of-the-art performance in self-supervised video action recognition on the UCF101 and HMDB51 datasets. Our method outperforms some methods that leverage the much larger-scale Kinetics dataset. These results demonstrate the efficacy of our proposed method to predict video transformations. We suggest our model as a powerful feature extractor for other tasks.

{\small
\bibliographystyle{ieee}
\bibliography{refs}

\begin{thebibliography}{10}\itemsep=-1pt

\bibitem{ahsan2019video}
U.~Ahsan, R.~Madhok, and I.~Essa.
\newblock Video jigsaw: Unsupervised learning of spatiotemporal context for
  video action recognition.
\newblock In {\em WACV}, pages 179--189. IEEE, 2019.

\bibitem{piergiovanni2019evolving}
P.~AJ, A.~Angelova, A.~Toshev, and M.~S. Ryoo.
\newblock Evolving space-time neural architectures for videos.
\newblock In {\em Proceedings of the IEEE international conference on computer
  vision}, pages 1793--1802, 2019.

\bibitem{buchler2018improving}
U.~Buchler, B.~Brattoli, and B.~Ommer.
\newblock Improving spatiotemporal self-supervision by deep reinforcement
  learning.
\newblock In {\em ECCV}, pages 770--786, 2018.

\bibitem{carreira2018short}
J.~Carreira, E.~Noland, A.~Banki-Horvath, C.~Hillier, and A.~Zisserman.
\newblock A short note about kinetics-600.
\newblock {\em arXiv preprint arXiv:1808.01340}, 2018.

\bibitem{i3d_2017}
J.~Carreira and A.~Zisserman.
\newblock Quo vadis, action recognition? a new model and the kinetics dataset.
\newblock In {\em CVPR}, pages 6299--6308, 2017.

\bibitem{dalal2006human}
N.~Dalal, B.~Triggs, and C.~Schmid.
\newblock Human detection using oriented histograms of flow and appearance.
\newblock In {\em ECCV}, pages 428--441. Springer, 2006.

\bibitem{deng2009imagenet}
J.~Deng, W.~Dong, R.~Socher, L.-J. Li, K.~Li, and L.~Fei-Fei.
\newblock Imagenet: A large-scale hierarchical image database.
\newblock In {\em 2009 IEEE conference on computer vision and pattern
  recognition}, pages 248--255. Ieee, 2009.

\bibitem{doersch2017multi}
C.~Doersch and A.~Zisserman.
\newblock Multi-task self-supervised visual learning.
\newblock In {\em CVPR}, pages 2051--2060, 2017.

\bibitem{el2019skip}
A.~El-Nouby, S.~Zhai, G.~W. Taylor, and J.~M. Susskind.
\newblock Skip-clip: Self-supervised spatiotemporal representation learning by
  future clip order ranking.
\newblock {\em arXiv preprint arXiv:1910.12770}, 2019.

\bibitem{fernando2017self}
B.~Fernando, H.~Bilen, E.~Gavves, and S.~Gould.
\newblock Self-supervised video representation learning with odd-one-out
  networks.
\newblock In {\em CVPR}, pages 3636--3645, 2017.

\bibitem{gan2018geometry}
C.~Gan, B.~Gong, K.~Liu, H.~Su, and L.~J. Guibas.
\newblock Geometry guided convolutional neural networks for self-supervised
  video representation learning.
\newblock In {\em CVPR}, pages 5589--5597, 2018.

\bibitem{han2019video}
T.~Han, W.~Xie, and A.~Zisserman.
\newblock Video representation learning by dense predictive coding.
\newblock In {\em CVPRW}, pages 0--0, 2019.

\bibitem{resnet3D_50}
K.~Hara, H.~Kataoka, and Y.~Satoh.
\newblock Can spatiotemporal 3d cnns retrace the history of 2d cnns and
  imagenet?
\newblock In {\em CVPR}, pages 6546--6555, 2018.

\bibitem{he2016deep}
K.~He, X.~Zhang, S.~Ren, and J.~Sun.
\newblock Deep residual learning for image recognition.
\newblock In {\em Proceedings of the IEEE conference on computer vision and
  pattern recognition}, pages 770--778, 2016.

\bibitem{jiang2019stm}
B.~Jiang, M.~Wang, W.~Gan, W.~Wu, and J.~Yan.
\newblock Stm: Spatiotemporal and motion encoding for action recognition.
\newblock In {\em Proceedings of the IEEE International Conference on Computer
  Vision}, pages 2000--2009, 2019.

\bibitem{kim2019self}
D.~Kim, D.~Cho, and I.~S. Kweon.
\newblock Self-supervised video representation learning with space-time cubic
  puzzles.
\newblock In {\em AAAI}, volume~33, pages 8545--8552, 2019.

\bibitem{klaser2008spatio}
A.~Kläser, M.~Marszalek, and C.~Schmid.
\newblock A spatio-temporal descriptor based on 3d-gradients.
\newblock In {\em BMCV}, 09 2008.

\bibitem{hmdb51}
H.~Kuehne, H.~Jhuang, E.~Garrote, T.~Poggio, and T.~Serre.
\newblock Hmdb: a large video database for human motion recognition.
\newblock In {\em ICCV}, pages 2556--2563. IEEE, 2011.

\bibitem{lee2017unsupervised}
H.-Y. Lee, J.-B. Huang, M.~Singh, and M.-H. Yang.
\newblock Unsupervised representation learning by sorting sequences.
\newblock In {\em CVPR}, pages 667--676, 2017.

\bibitem{misra2016shuffle}
I.~Misra, C.~L. Zitnick, and M.~Hebert.
\newblock Shuffle and learn: unsupervised learning using temporal order
  verification.
\newblock In {\em ECCV}, pages 527--544. Springer, 2016.

\bibitem{ravanelli2020multi}
M.~Ravanelli, J.~Zhong, S.~Pascual, P.~Swietojanski, J.~Monteiro, J.~Trmal, and
  Y.~Bengio.
\newblock Multi-task self-supervised learning for robust speech recognition.
\newblock In {\em ICASSP}, pages 6989--6993. IEEE, 2020.

\bibitem{sarkar2020self}
P.~Sarkar and A.~Etemad.
\newblock Self-supervised learning for ecg-based emotion recognition.
\newblock In {\em ICASSP}, pages 3217--3221. IEEE, 2020.

\bibitem{CrossLearn}
N.~Sayed, B.~Brattoli, and B.~Ommer.
\newblock Cross and learn: Cross-modal self-supervision.
\newblock In {\em Pattern Recognition}, pages 228--243, Cham, 2019. Springer
  International Publishing.

\bibitem{scovanner20073}
P.~Scovanner, S.~Ali, and M.~Shah.
\newblock A 3-dimensional sift descriptor and its application to action
  recognition.
\newblock In {\em Proceedings of the 15th ACM international conference on
  Multimedia}, pages 357--360, 2007.

\bibitem{ucf101}
K.~Soomro, A.~R. Zamir, and M.~Shah.
\newblock A dataset of 101 human action classes from videos in the wild.
\newblock {\em Center for Research in Computer Vision}, 2, 2012.

\bibitem{CMC}
Y.~Tian, D.~Krishnan, and P.~Isola.
\newblock Contrastive multiview coding, 2020.

\bibitem{tran2015learning}
D.~Tran, L.~Bourdev, R.~Fergus, L.~Torresani, and M.~Paluri.
\newblock Learning spatiotemporal features with 3d convolutional networks.
\newblock In {\em CVPR}, pages 4489--4497, 2015.

\bibitem{villegas2017decomposing}
R.~Villegas, J.~Yang, S.~Hong, X.~Lin, and H.~Lee.
\newblock Decomposing motion and content for natural video sequence prediction.
\newblock {\em arXiv preprint arXiv:1706.08033}, 2017.

\bibitem{wang2013action}
H.~Wang and C.~Schmid.
\newblock Action recognition with improved trajectories.
\newblock In {\em CVPR}, pages 3551--3558, 2013.

\bibitem{wang2019self}
J.~Wang, J.~Jiao, L.~Bao, S.~He, Y.~Liu, and W.~Liu.
\newblock Self-supervised spatio-temporal representation learning for videos by
  predicting motion and appearance statistics.
\newblock In {\em CVPR}, pages 4006--4015, 2019.

\bibitem{willems2008efficient}
G.~Willems, T.~Tuytelaars, and L.~Van~Gool.
\newblock An efficient dense and scale-invariant spatio-temporal interest point
  detector.
\newblock In {\em ECCV}, pages 650--663. Springer, 2008.

\bibitem{xu2019self}
D.~Xu, J.~Xiao, Z.~Zhao, J.~Shao, D.~Xie, and Y.~Zhuang.
\newblock Self-supervised spatiotemporal learning via video clip order
  prediction.
\newblock In {\em CVPR}, pages 10334--10343, 2019.

\bibitem{zhang2016colorful}
R.~Zhang, P.~Isola, and A.~A. Efros.
\newblock Colorful image colorization.
\newblock In {\em ECCV}, pages 649--666. Springer, 2016.

\end{thebibliography}
}

\end{document}